\documentclass[sigconf]{acmart}

\usepackage{amsmath,amsfonts}
\usepackage{graphicx}
\usepackage{diagbox}
\usepackage{tcolorbox}
\usepackage{multirow}
\usepackage{booktabs}
\usepackage{makecell}
\usepackage{xcolor}
\usepackage{colortbl}
\usepackage{balance}

\copyrightyear{2026}
\acmYear{2026}
\setcopyright{cc}
\setcctype{by}
\acmConference[KDD '26]{Proceedings of the 32nd ACM SIGKDD Conference on Knowledge Discovery and Data Mining V.2}{August 09--13, 2026}{Jeju Island, Republic of Korea}
\acmBooktitle{Proceedings of the 32nd ACM SIGKDD Conference on Knowledge Discovery and Data Mining V.2 (KDD '26), August 09--13, 2026, Jeju Island, Republic of Korea}
\acmDOI{10.1145/3770855.3818873}
\acmISBN{979-8-4007-2259-2/2026/08}

\begin{document}

\title[TF-SNO: Time-Frequency Gated Spectral Neural Operators]{TF-SNO: Time-Frequency Gated Spectral Neural Operators for Learning Non-Stationary Partial Differential Equations}

\author{Yitian Zhou}
\affiliation{\institution{University of Electronic Science and Technology of China}
    \city{Chengdu}
    \country{China}}
\email{ytzhouuu@gmail.com}

\author{Chaoning Zhang}
\affiliation{\institution{University of Electronic Science and Technology of China}
    \city{Chengdu}
    \country{China}}
\email{chaoningzhang1990@gmail.com}

\author{Zhenzhen Huang}
\affiliation{\institution{University of Electronic Science and Technology of China}
    \city{Chengdu}
    \country{China}}
\email{alley10086@gmail.com}

\author{Haoxuan Yu}
\affiliation{\institution{University of Electronic Science and Technology of China}
    \city{Chengdu}
    \country{China}}
\email{2024090902011@std.uestc.edu.cn}

\author{Jiaquan Zhang}
\affiliation{\institution{University of Electronic Science and Technology of China}
    \city{Chengdu}
    \country{China}}
\email{jiaquanzhang2005@gmail.com}

\author{Yiran Li}
\affiliation{\institution{University of Electronic Science and Technology of China}
    \city{Chengdu}
    \country{China}}
\email{2025080908010@std.uestc.edu.cn}

\author{Fan Mo}
\affiliation{\institution{University of Electronic Science and Technology of China}
    \city{Chengdu}
    \country{China}}
\email{fan.mo@kellogg.ox.ac.uk}

\author{Kuien Liu}
\affiliation{\institution{Institute of Software, Chinese Academy of Sciences}
    \city{Beijing}
    \country{China}}
\email{kuien@iscas.ac.cn}

\author{Jie Zou}
\affiliation{\institution{University of Electronic Science and Technology of China}
    \city{Chengdu}
    \country{China}}
\email{zoujie0806@gmail.com}

\author{Caiyan Qin}
\authornote{Corresponding author, e-mail: qincaiyan@hit.edu.cn}
\affiliation{\institution{Harbin Institute of Technology}
    \city{Shenzhen}
    \country{China}}
\email{qincaiyan@hit.edu.cn}

\author{Yang Yang}
\affiliation{\institution{University of Electronic Science and Technology of China}
    \city{Chengdu}
    \country{China}}
\email{yang.yang@uestc.edu.cn}

\renewcommand{\shortauthors}{Yitian Zhou et al.}

\begin{abstract}
Non-stationary partial differential equations (PDEs) arise throughout scientific computing, where the dominant frequency content and energy distribution can drift over time. While efficient in PDE solving, many spectral neural operators apply a shared spectral response across rollout stages, leading to mismatch with time-varying spectra in non-stationary systems. To address this issue, we propose Time-Frequency Gated Spectral Neural Operator (TF-SNO), a state-adaptive framework with learnable time-frequency gating inside spectral blocks. TF-SNO extracts compact frequency-domain and physical-space statistics from the current state to generate modulation coefficients, enabling the spectral response to evolve with the dynamics. TF-SNO learns temporal variation implicitly from the evolving state without introducing an explicit time dimension or time embedding, keeping the modeling complexity low. We further embed the adaptive operator blocks to accurately capture the multi-scale features, thereby improving long-horizon stability. Experiments on six non-stationary PDE benchmarks in 1D and 2D demonstrate that TF-SNO significantly reduces prediction errors and improves robustness compared to strong baselines, with particularly clear gains in long rollout, suggesting the effectiveness of state-dependent spectral adaptation in modeling non-stationary physical systems.
\end{abstract}



\maketitle

\section{Introduction}

Many scientific systems are governed by time-dependent partial differential equations (PDEs) with time-varying spectral energy distribution, including turbulence, wave propagation, and multiscale transport in materials. In such non-stationary PDEs, energy transfers across frequencies as the system evolves, with switching dominant modes and short-lived high-frequency bursts, resulting in time-varying spectra. For example, in unsteady turbulence, energy transfer across scales can vary over time rather than staying in a single steady cascade pattern \cite{bos2024unsteady}.
Accurate simulation of these dynamics often requires fine spatial and temporal resolution, especially when sharp gradients, intermittent events, or multiscale interactions are present. While classical numerical solvers, such as finite difference, finite volume, finite element, and spectral methods, can be highly accurate, repeated simulations across many initial conditions, parameters, or control inputs remain computationally expensive \cite{leveque2002finite}. This cost is a major bottleneck in scientific workflows like uncertainty quantification, inverse problems, and design loops. In response, \emph{operator learning} has been motivated to learn mappings between function spaces that generalize across input fields, rather than fitting a solution to a fixed setup, which is also consistent with broader efforts in predictive representation learning \cite{202605.0435,zhou2026similarity,zhang2026lightweight}.

\begin{figure}[t]
    \centering
    \includegraphics[width=\linewidth]{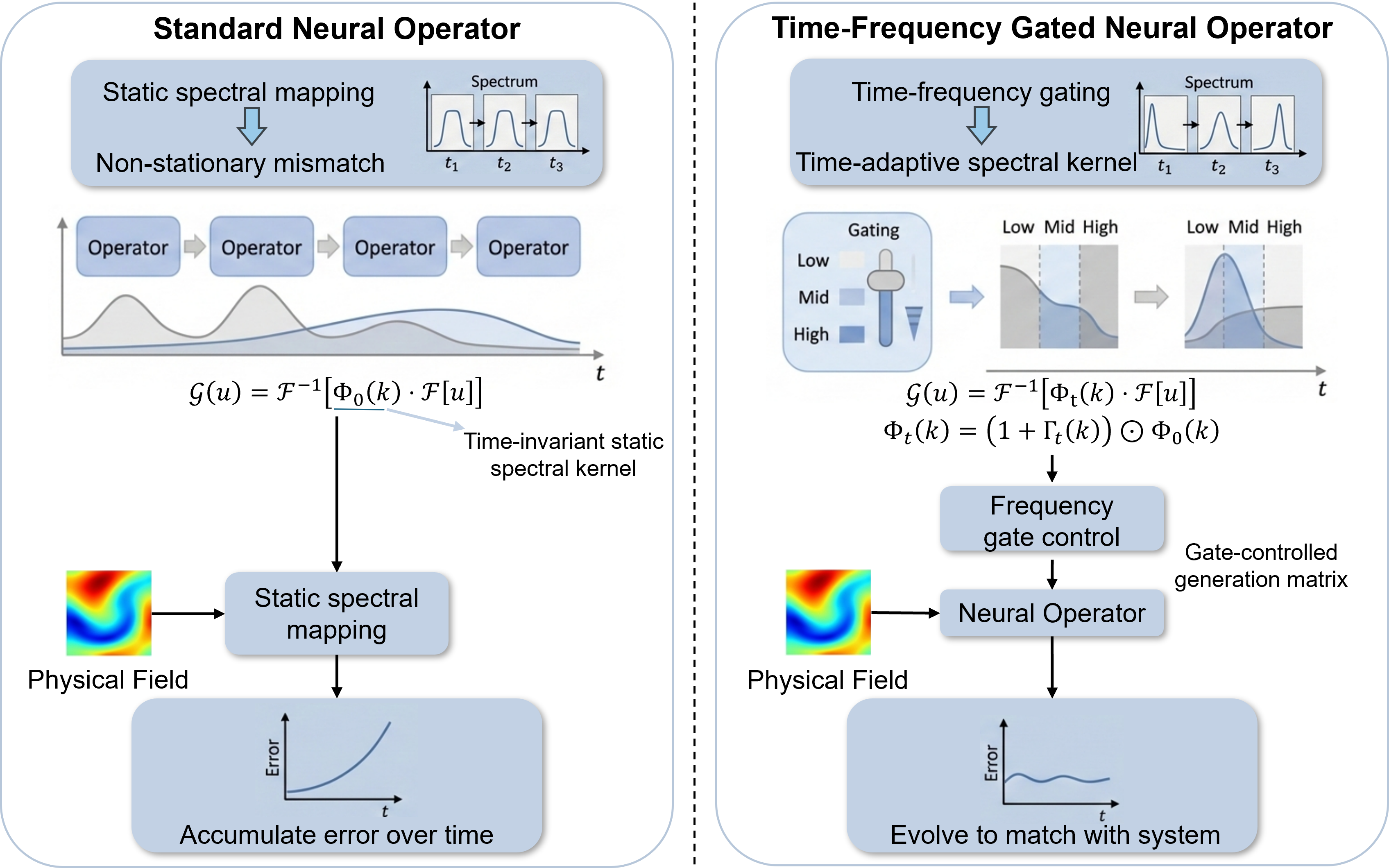}
    \caption{Concept of state-dependent gating compared with static spectral kernel in standard neural operators.}
    \label{fig:concept}
\end{figure}

Following this, neural operators have recently emerged as a promising approach for fast PDE surrogates because they can generalize across initial conditions, forcing terms, or coefficients with function mappings \cite{kovachki2023neural}. Among them, the Fourier Neural Operator (FNO) and its spectral variants are widely adopted due to FFT-based efficiency and strong generalization to discretization \cite{li2021fourier,tran2023factorized,xiao2024orthogonal,li2023geofno}. FNO performs well on stationary or quasi-stationary PDEs, where a shared global spectral transform can represent the dominant modes consistently. As shown in Fig. \ref{fig:concept}, however, its core spectral mapping is typically parameterized by Fourier weights that are shared across samples and reused at every evaluation step, applying the same spectral response throughout the rollout. This does not mean that FNO is globally static. Stacked nonlinear layers, residual connections, and pointwise transforms still make the whole network input-dependent. The precise limitation is that the learned Fourier kernel does not explicitly perform state-conditioned band routing.
For non-stationary dynamics, where spectral energy migrates across modes, and transient structures appear at different stages, this fixed-kernel design can cause spectral mismatch and accumulated rollout error \cite{gao2025dynamic,cao2024laplace,tiwari2025cono}. 

A common response in prior work is to introduce \emph{explicit temporal modeling} in the operator architecture, for example by treating time as an additional input coordinate, adding a temporal branch, or learning spatio-temporal operators over space-time domains to improve temporal generalization \cite{lu2021learning,diab2025temporal,nayak2025tideeponet,cao2024spectralrefiner}. While effective in many settings, these designs typically increase modeling and implementation complexity by expanding the input domain or augmenting network components dedicated to time \cite{chen2025neural}. As shown in Fig. \ref{fig:concept}, they lack an explicit mechanism for \emph{state-dependent} adaptation of spectral responses as the evolving system transitions between phases with different dominant frequency content.
For autonomous non-stationary PDEs, the evolutionary stage is often encoded in the current state through its amplitude, roughness, and spectral energy distribution. This observation suggests a lighter design that retains FFT-level efficiency but allows the spectral kernel to adapt its band-wise response according to the current hidden state.

To address this, we propose \textbf{Time-Frequency Gated Spectral Neural Operator (\textit{TF-SNO})}, which introduces learnable time-frequency gating within spectral operator blocks. \textit{TF-SNO} extracts compact frequency-domain statistics and time-varying physical-space statistics from the current state, uses a lightweight gating network to generate frequency band-wise modulation coefficients, and dynamically reweights Fourier modes accordingly. Notably, TF-SNO does not require introducing time as an explicit input dimension or time embedding. Instead, temporal variation is modeled implicitly through the evolving state, keeping the operator form and modeling complexity close to standard FFT-based neural operators with static spectral kernels. Moreover, we embed the adaptive spectral operator blocks into a U-shaped multiscale backbone so that state-adaptive spectral modulation is applied consistently across resolutions, enabling the model to represent both coarse global structures and fine transient details. For training, we combine supervised data loss with a global H$^1$ term to constrain gradient behavior to improve physical consistency and long-horizon stability.

The main contributions of this paper are summarized as follows:
\begin{itemize}
\item We propose \textit{TF-SNO}, a state-adaptive time-frequency gated neural operator that enables frequency band-wise modulation of spectral responses during rollouts, addressing non-stationary dynamics with evolving spectra without explicit time conditioning.
\item We integrate the gated spectral operator blocks into a U-shaped multiscale architecture, allowing state-adaptive spectral modulation to operate coherently across resolutions and improving representation of coupled global structures and transient local features.
\item We adopt an H$^1$-regularized training objective, and empirically demonstrate improved gradient fidelity and rollout stability on non-stationary PDE benchmarks.
\end{itemize}

\section{Related Work}
\subsection{Traditional Numerical PDE Solvers}
Classical PDE solvers remain the standard tool for high-fidelity simulation in science and engineering. Finite difference, finite volume, finite element, and spectral methods provide systematic control of accuracy and stability, with finite volume methods being especially effective for conservation laws and transport-dominated systems \cite{leveque2002finite}. However, repeated high-resolution solves across many initial conditions, parameters, or control settings remain costly in inverse problems, uncertainty quantification, and design loops, motivating surrogate models that approximate solution operators rather than solving each instance from scratch \cite{kovachki2023neural}.

\subsection{Physics-Informed Neural Networks}
Physics-Informed Neural Networks (PINNs) incorporate governing equations into training by penalizing PDE residuals and boundary or initial conditions \cite{raissi2019physics,huang2026rethinking}. Such residual supervision can reduce the need for dense labels and improve physical consistency, although optimization pathologies and failure modes can arise in challenging PDE regimes \cite{wang2021understanding}. PINN ideas also influence sequential and time-marching neural solvers for long-horizon prediction \cite{he2024sequential}. Meanwhile, PINNs typically operate at the level of approximating a single solution field for a given setup, while many scientific tasks require generalization across varying inputs, motivating operator-learning approaches that can combine data supervision with physics constraints \cite{white2023physics,li2024physics,eshaghi2025vino}.

\subsection{Operator Learning and Neural Operators}
Operator learning aims to learn mappings between function spaces and has become a central framework for PDE surrogate modeling \cite{kovachki2023neural}. Inspired by the universal approximation theorem for operators \cite{chen1995universal}, DeepONet provides a general framework for approximating nonlinear operators with branch and trunk networks \cite{lu2021learning}.
Building on this foundation, Neural Operators have further advanced the field toward large-scale applications. Fourier and transform-domain neural operators achieve efficient global mixing and strong resolution generalization through compact spectral or transformed representations \cite{li2021fourier,tran2023factorized,xiao2024orthogonal,cao2024laplace}. Beyond structured grids, graph-based operator methods support irregular meshes and complex geometries by expressing kernel integration through message passing, kernel approximation, or geometric constraints \cite{li2020neural,li2020multipole,yin2024dimon,zhang2026geometric,hu2024better}. Wavelet-based operators improve locality and multiresolution modeling for transient structures and generative PDE simulation \cite{hu2025wavelet,tripura2023wavelet}, while U-Net-like neural operators strengthen feature fusion across scales through encoder-decoder pathways and skip connections \cite{li2023iufno,rahman2023uno}. 
Physics-informed neural operators (PINO) further integrate PDE residual constraints into operator learning, aiming to improve physical consistency and data efficiency \cite{li2024physics,eshaghi2025vino}, whereas localized-kernel and finite-difference-inspired operators emphasize stencil-like updates for stable rollouts \cite{cheng2025pde,liu2024localized}.

For time-dependent PDEs, existing methods often introduce time as a coordinate, a temporal branch, or a spatio-temporal trajectory representation \cite{diab2025temporal,lu2021learning,nayak2025tideeponet}. Recent spatio-temporal, memory-based, pretrained, and autoregression-free operators further learn trajectory-level mappings to reduce step-by-step error accumulation \cite{cao2024spectralrefiner,zhang2026autoregression,buitrago2025benefits}.
TF-SNO differs by retaining an FFT-based operator form while adding explicit state-conditioned band-wise modulation inside spectral blocks, targeting autonomous non-stationary PDEs whose evolutionary stage is reflected in the current state.

\begin{figure*}[t]
    \centering
    \includegraphics[width=0.86\linewidth]{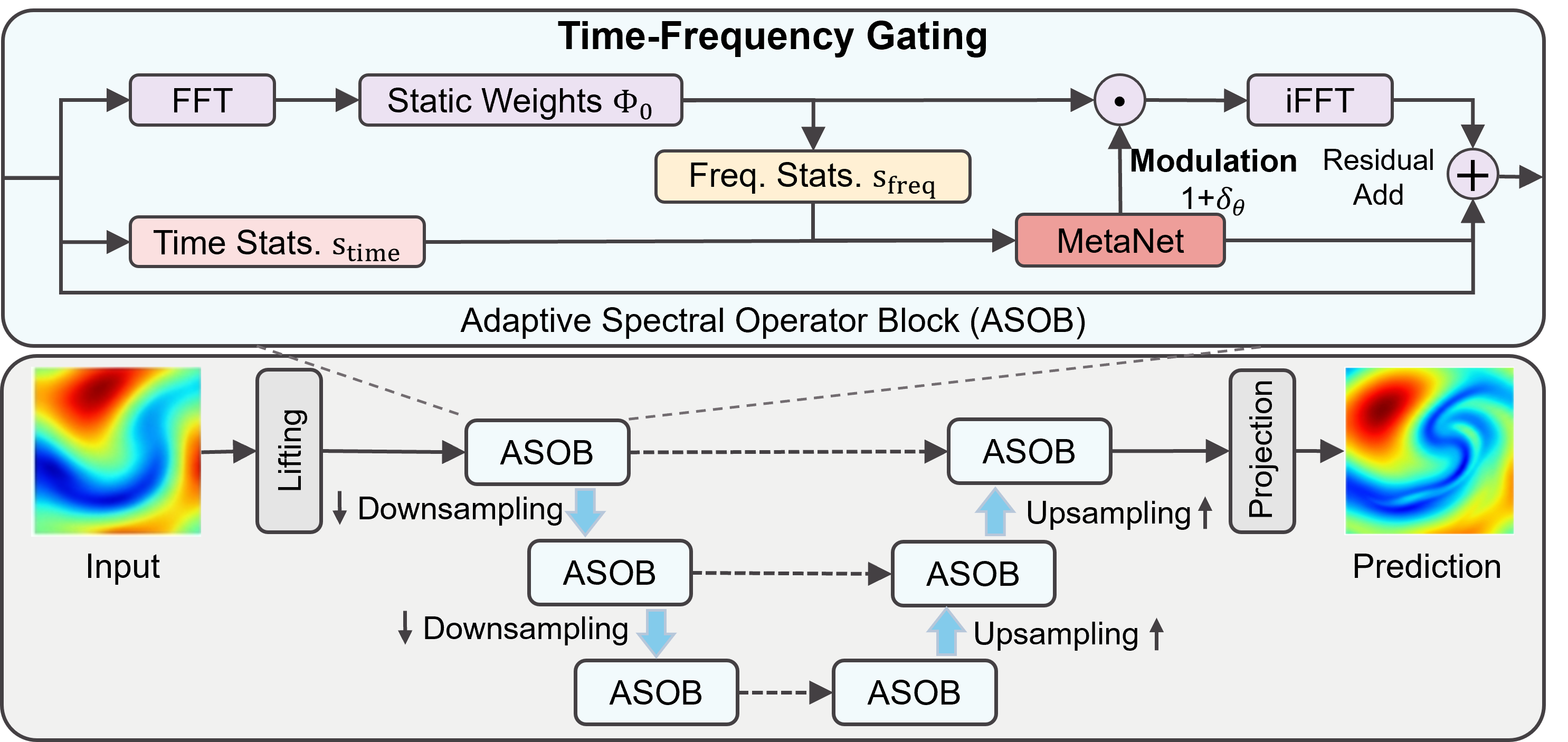}
    \caption{Architecture of the TF-SNO. TF-SNO introduces time-frequency gating in the Adaptive Spectral Operator Block (ASOB) and embeds ASOBs into a multiscale U-shaped backbone.}
    \label{fig:main}
\end{figure*}

\section{Method}
In this section, we systematically describe the overall structure and training details of the proposed \textit{TF-SNO} framework with complexity and local stability analysis. 

\subsection{Problem Formulation}

Non-stationary PDE dynamics are characterized by the evolving spectral content of the solution, which can change substantially across different stages of the trajectory. This stage variation is often {implicitly encoded} in the current system state, with different dominant frequencies and energy distribution patterns. 
We consider a time-dependent PDE on a spatial domain $\Omega$, with state $u(x,t)$ evolving under a (generally nonlinear) evolution rule. Under a fixed time step $\Delta t$, the one-step mapping in discrete form is:
\begin{equation}
    u(\cdot,t_{n+1}) = \mathcal{T}\big(u(\cdot,t_n)\big),
\end{equation}
where $\mathcal{T}$ denotes the ground-truth operator for one-step advancement.
Neural Operators aim to learn such mappings in functional spaces, enabling generalization across input functions rather than fitting discrete point-wise values for a single trajectory. Many FFT-based neural operators parameterize a translation-invariant spectral convolution as:
\begin{equation}
    \mathcal{G}_{\theta}(u) = \mathcal{F}^{-1}\!\left[ \Phi_{\theta}(k)\cdot \mathcal{F}[u] \right],
\end{equation}
where $\Phi_{\theta}(k)$ is a learned {fixed} spectral kernel shared across all rollout stages, and $\mathcal{F}$ (resp. $\mathcal{F}^{-1}$) denotes the spatial Fourier transform (resp. its inverse). This form is computationally efficient, but its single static spectral response can be restrictive when the dominant frequency content changes across trajectory stages.
Our goal is to learn an evolution surrogate that can be rolled out over long horizons. Given training trajectories $\{u(\cdot,t_n)\}_{n=0}^{N}$ sampled from diverse initial states and parameter settings, we learn an autoregressive operator $\mathcal{G}_\theta$:
\begin{equation}
    \hat{u}(\cdot,t_{n+1}) = \mathcal{G}_\theta\big(\hat{u}(\cdot,t_n)\big), \quad \hat{u}(\cdot,t_0)=u(\cdot,t_0),
\end{equation}
where $\hat{u}$ denotes the predicted rollout state. 
The proposed state-only formulation assumes that the next state is determined by the current state, which is appropriate for autonomous PDEs. When the system is non-autonomous, similar spatial states may evolve differently under different forcing phases. In such cases, the gate can be extended by concatenating an explicit time encoding to the state statistics.

\subsection{TF-SNO: Time-Frequency Gated Spectral Neural Operator}
In non-stationary regimes, the spectral characteristics can vary markedly across different {trajectory stages}, where high-energy disturbance phases tend to exhibit stronger high-frequency components, while calmer phases are often dominated by low-frequency modes. 
As illustrated in Fig.~\ref{fig:main}, we propose \textit{TF-SNO}, which introduces a state-dependent time-frequency gating mechanism to endow FFT-based neural operators with adaptive spectral responses. Instead of relying on a single static spectral kernel throughout the rollout, TF-SNO modulates the spectral response conditioned on compact statistics extracted from the current hidden state, so that different stages can activate different frequency modes.

We parameterize each spectral operator block with a basis spectral response $\Phi_0(k)$ and a state-dependent gating function:
\begin{equation}
    \Phi(k;\,z) = \bigl(1+\delta_\theta(k;\,z)\bigr)\odot \Phi_0(k),
\end{equation}
where $z$ denotes the current hidden representation within the block. The basis kernel $\Phi_0(k)$ captures global and invariant frequency-domain features, while $\delta_\theta(k;\,z)$ dynamically adjusts the relative importance of different frequency modes based on the current state.

From a signal processing perspective, $\delta_\theta$ acts as an adaptive filter in the spectral domain. By allowing the filter response to vary with the evolving state, TF-SNO can capture non-stationary behaviors such as frequency-domain energy migration, mode switching, and transient bursts. We implement this mechanism in an \textit{Adaptive Spectral Operator Block (ASOB)}:
\begin{equation}
    \mathrm{ASOB}_{\theta(z)}=\mathcal{F}^{-1}[\Phi(k;z)\odot \mathcal{F}[z]] + \mathcal{W}(z),
\end{equation}
which preserves the computational efficiency of FFT-based operators while enabling state-adaptive spectral plasticity.

\subsubsection{Time-Frequency Domain Statistics}
Let $z\in\mathbb{R}^{C\times N}$ denote the hidden representation entering an ASOB, where $C$ is the channel dimension and $N$ is the spatial resolution at the current scale. We first compute the truncated Fourier representation $\hat{z}_{\mathrm{ft}}$ and obtain the standard spectral convolution output:
\begin{equation}\label{eq:fft}
    \hat{y}_{\mathrm{ft}}^{(0)} = \Phi_0(k)\cdot \hat{z}_{\mathrm{ft}}.
\end{equation}
At each gated spectral block, TF-SNO extracts compact statistics from the current state to drive spectral modulation.

\paragraph{Frequency-domain statistics}
Let $k\in\mathbb{Z}^d$ denote a multi-dimensional Fourier frequency index, and let $\mathcal{K}\subset\mathbb{Z}^d$ be the truncated set of retained modes used by the spectral operator block. We denote $|\mathcal{K}|$ as the number of retained modes.
For the truncated spectral representation within the block, we compute:
\begin{equation}
 s_{\mathrm{freq}} = \bigl[\log E, \mu_{\mathcal{K}}, \sigma_{\mathcal{K}}, \mathrm{Skew}_{\mathcal{K}} \bigr]\in\mathbb{R}^4,
\end{equation}
where $E$ is spectral energy over $\mathcal{K}$, $\mu_{\mathcal{K}}$ is the (power-weighted) spectral centroid, $\sigma_{\mathcal{K}}$ is spectral spread, and $\mathrm{Skew}_{\mathcal{K}}$ is spectral skewness, summarizing how energy is distributed over the retained multi-dimensional modes. For $d>1$, we compute these summary statistics using the radial frequency magnitude $\|k\|$, so that the feature dimension remains constant across spatial dimensions.

\paragraph{Physical-space statistics}
From the current state, we compute:
\begin{equation}
    s_{\mathrm{time}} = \bigl[\log \sigma_z, R_{\mathrm{rough}}, \kappa_z, c\bigr]\in\mathbb{R}^4,
\end{equation}
where $\log \sigma_z$ characterizes the amplitude scale, $R_{\mathrm{rough}}$ measures local roughness, $\kappa_z$ is kurtosis, and $c$ is a shape descriptor capturing transient spatial characteristics.

These two statistics are concatenated to form a joint embedding:
\begin{equation}
    s_{\mathrm{tf}} = [s_{\mathrm{freq}}, s_{\mathrm{time}}]\in\mathbb{R}^8.
\end{equation}
This 8D vector is not a compressed replacement for the hidden state. It controls only the gating network, while the full hidden representation remains available to the spectral and residual path, making it compact without forcing the model to discard spatial information.

\subsubsection{Gating Network and Spectral Modulation}
We use a lightweight gating network (MetaNet) $F_{\text{meta}}$ to map the joint embedding $s_{\mathrm{tf}}$ to mode-wise coefficients over the retained frequency set $\mathcal{K}$. The gating tensor is computed as:
\begin{equation}\label{eq:meta}
    g_\theta (z) = \tanh\left(F_{\text{meta}}(s_{\mathrm{tf}}(z))\right)\in\mathbb{R}^{C\times|\mathcal{K}|}.
\end{equation}
A learnable gain factor $\alpha$ is used to control the modulation strength:
\begin{equation}
    \delta_\theta (z) = \alpha\cdot g_\theta (z),\quad \alpha=\mathrm{softplus}(\alpha^*),
\end{equation}
where $\alpha^*$ is an unconstrained scalar parameter.

Given the standard spectral convolution output $\hat{y}_{\mathrm{ft}}^{(0)}$ in Eq.~\eqref{eq:fft}, TF-SNO applies state-dependent modulation as: 
\begin{equation}
    \hat{y}_{\mathrm{ft}} = \hat{y}_{\mathrm{ft}}^{(0)}\odot\bigl(1+\delta_\theta(z)\bigr),
\end{equation}
where the elementwise product is applied over $(c,k)$ for $k\in\mathcal{K}$. Since $\delta_\theta$ is recomputed at each ASOB, different trajectory stages can induce different frequency reweighting patterns without requiring explicit time conditioning.

\subsubsection{Adaptive Spectral Operator Block with Residual Path}
Each ASOB combines a modulated spectral path with a local residual path. The block output is
\begin{equation}
    z'=\mathrm{GELU}\left(\mathcal{F}^{-1}[\hat{y}_{\mathrm{ft}}] + \mathcal{W}(z)\right),
\end{equation}
where $\mathcal{W}$ is a $1\times1$ convolution in the physical space. This design preserves the efficiency of FFT-based operators while enabling state-adaptive spectral responses via the gating mechanism.

\subsection{Multiscale TF-SNO Backbone}
Non-stationary PDE solutions often exhibit both global trends and localized transient structures, with dominant scales and frequency content varying across trajectory stages. To capture such mixed-scale features, TF-SNO embeds the proposed Adaptive Spectral Operator Blocks (ASOBs) into a U-shaped multiscale backbone inspired by U-NO \cite{rahman2023uno} as shown in Fig. \ref{fig:main}. 
The encoder reduces spatial resolution to enlarge the receptive field and emphasize slowly varying components, while the decoder restores resolution to reconstruct fine structures. Skip connections link matching levels to preserve high-resolution information and support feature fusion across resolutions.
Applying ASOBs at multiple resolutions allows coarse blocks to stabilize low-frequency trends, while fine blocks amplify or suppress high-frequency content when transient structures appear. This enables the same state-adaptive gating principle to operate at different spatial scales.

\subsection{Training Objectives}
Given paired one-step training samples $(u_t, u_{t+1})$, TF-SNO predicts the next state $\hat{u}_{t+1}$ from $u_t$. The supervised objective contains a field-value error and a gradient-matching error:
\begin{equation}
    \mathcal{L}_{\text{sup}} = \| \hat{u}_{t+1} - u_{t+1} \|_2^2 + \lambda_{H^1} \| \nabla_h \hat{u}_{t+1} - \nabla_h u_{t+1} \|_2^2.
\end{equation}
where $\nabla_h$ denotes the finite-difference spatial gradient and $\lambda_{H^1}$ controls the weight of the gradient-matching term. 
For 1D equations, the derivative is computed along the spatial axis. For 2D equations, derivatives along both spatial directions are included. Interior points use central differences, while boundary points are handled according to the benchmark boundary condition, with periodic padding for periodic domains. 
This H$^1$-based regularization penalizes gradient mismatch, which encourages fidelity not only in field values but also in local spatial structures. 

\subsection{Computational Complexity}

Compared with a standard FFT-based spectral operator with spectral convolution, TF-SNO introduces extra computation mainly from the gating network $F_{\text{meta}}$ and the extraction of low-dimensional time- and frequency-domain statistics.
As in Eq.~\eqref{eq:meta}, the gating produces a modulation tensor of size $C\times|\mathcal{K}|$, yielding an additional per-block cost on the order of $O(C|\mathcal{K}|)$ for elementwise modulation, plus the cost of the gating network $F_{\text{meta}}$.
It is linear in both $C$ and $|\mathcal{K}|$, typically small relative to the dominant costs in spectral operator blocks, such as FFTs and convolutions in spectral space.

\subsection{Local Stability Analysis}
We provide a local stability analysis for a single ASOB to characterize the additional sensitivity term introduced by state-dependent gating. On a fixed input $z$, $\delta(z)=\alpha\tanh(F_{meta}(s(z)))$, ASOB can be written as:
\begin{equation}
A(z)=\sigma\big(\mathcal{F}^{-1}\big( ((1+\delta(z))\odot \Phi_0)\odot\mathcal{F}z\big)+Wz\big).
\end{equation} 
Considering gate tensors $\delta_a$ and $\delta_b$ and their block outputs $A_a(z)$ and $A_b(z)$, under unitary FFT normalization, assuming $\sigma$ is $L_{\sigma}$-Lipschitz and $M_\Phi=\|\Phi_0\|_{\infty}$, we have:
\begin{equation}
\|A_a(z)-A_b(z)\|_2\le L_\sigma M_\Phi \|z\|_2\, \|\delta_a-\delta_b\|_\infty.
\end{equation}
Thus, the variation caused by gate perturbation is explicitly bounded and proportional to the gate difference, so adaptive gating does not introduce uncontrolled sensitivity.

Considering the dependence of $\delta(z)$ on $z$, assume $s(z)$ is $L_s$-Lipschitz on $\mathcal{B}_R=\{z:\|z\|_2\le R\}$ and $F_{\text{meta}}$ is $L_m$-Lipschitz. Since $\tanh$ is 1-Lipschitz:
\begin{equation}
\|\delta(z_1)-\delta(z_2)\|_\infty\le\alpha L_m L_s \|z_1-z_2\|_2.
\end{equation}
For $z_1,z_2\in\mathcal{B}_R$,
\begin{equation}
\|A(z_1)-A(z_2)\|_2 \le L_\sigma \big[ (1+\alpha)M_\Phi
+M_W+\alpha M_\Phi L_m L_s R \big]\, \|z_1-z_2\|_2.
\end{equation}
The bracketed constant decomposes the amplification into the static spectral term, the residual term, and the state-dependent term induced by gating. Therefore, the gating mechanism remains analyzable with a controllable additive sensitivity factor. For a depth-$D$ network, enforcing the bracketed constant below one through spectral normalization on $\Phi_0$ and $W$ or by bounding $\alpha$ yields a contractive composite operator in the analyzed region.

This analysis also gives a Koopman-style interpretation. TF-SNO does not claim to learn a single global Koopman operator. Each ASOB instantiates a state-conditioned linear operator:
\begin{equation}
K_z=\mathcal{F}^{-1}\ \big(\mathrm{Diag}((1+\delta(z))\odot\Phi_0)\big)\mathcal{F}+W.
\end{equation}
Thus, TF-SNO can be interpreted as learning a family of local operators indexed by the current state, dynamically adjusting the spectral representation according to the current dynamical regime.

\section{Experiments}

\begin{table*}[t]
\centering
\caption{Rollout results of all comparison methods at prediction step of \{1,10,25,50\} for six benchmarks.}
\label{tab:rollout_horizons}
\setlength{\tabcolsep}{2.5pt}
\begin{tabular}{ll|ccc|ccc|ccc|ccc}
\toprule
\multirow{2}{*}{Eq.} & \multirow{2}{*}{Model} & \multicolumn{3}{c|}{Step=1} & \multicolumn{3}{c|}{Step=10} & \multicolumn{3}{c|}{Step=25} & \multicolumn{3}{c}{Step=50} \\
\cmidrule(lr){3-5}\cmidrule(lr){6-8}\cmidrule(lr){9-11}\cmidrule(lr){12-14}
 &  & MSE & L2 & H1 & MSE & L2 & H1 & MSE & L2 & H1 & MSE & L2 & H1 \\
\midrule
\multirow{7}{*}{E1}
& TF-SNO   & \cellcolor{gray!25}\textbf{0.00000} & \cellcolor{gray!25}\textbf{0.00205} & \cellcolor{gray!25}\textbf{0.00255} & \cellcolor{gray!25}\textbf{0.00004} & \cellcolor{gray!25}\textbf{0.01504} & \cellcolor{gray!25}\textbf{0.02164} & \cellcolor{gray!25}\textbf{0.00019} & \cellcolor{gray!25}\textbf{0.03262} & 0.04856 & \cellcolor{gray!25}\textbf{0.00075} & \cellcolor{gray!25}\textbf{0.06513} & 0.09712 \\
& FNO      & 0.00001 & 0.00760 & 0.01140 & 0.00064 & 0.05380 & 0.06364 & 0.00220 & 0.10314 & 0.12024 & 0.00790 & 0.20031 & 0.22252 \\
& DeepONet & 0.00244 & 0.12050 & 0.20571 & 0.00155 & 0.09539 & 0.16611 & 0.00032 & 0.04363 & 0.07515 & 0.00220 & 0.11094 & 0.24780 \\
& U-NO     & 0.00000 & 0.00224 & 0.00385 & 0.00005 & 0.01746 & 0.02389 & 0.00023 & 0.03768 & \cellcolor{gray!25}\textbf{0.04455} & 0.00096 & 0.07829 & \cellcolor{gray!25}\textbf{0.07815} \\
& PINO     & 0.00001 & 0.00702 & 0.01286 & 0.00084 & 0.06014 & 0.08642 & 0.00480 & 0.14457 & 0.20067 & 0.02650 & 0.33845 & 0.45406 \\
& WNO      & 0.00001 & 0.00652 & 0.00699 & 0.00069 & 0.06209 & 0.06242 & 0.00384 & 0.14712 & 0.14593 & 0.02644 & 0.35840 & 0.42546 \\
& FINO     & 0.00000 & 0.00243 & 0.00384 & 0.00011 & 0.02251 & 0.03204 & 0.00054 & 0.05158 & 0.06548 & 0.00211 & 0.10413 & 0.13842 \\
\midrule
\multirow{7}{*}{E2}
& TF-SNO   & {0.00001} & 0.06190 & 0.03742 & {0.00037} & 0.39929 & 0.28825 & {0.00053} & 0.41901 & 0.41712 & \cellcolor{gray!25}\textbf{0.00058} & \cellcolor{gray!25}\textbf{0.28102} & 0.33864 \\
& FNO      & 0.00002 & 0.07528 & 0.05260 & 0.00074 & 0.56479 & 0.47401 & 0.00196 & 0.80795 & 0.89714 & 0.00534 & 0.85203 & 1.05332 \\
& DeepONet & 0.00309 & 0.99210 & 0.97742 & 0.18274 & 7.35459 & 2.67387 & 0.28402 & 9.69643 & 4.71948 & 0.26564 & 6.04704 & 3.86436 \\
& U-NO     & \cellcolor{gray!25}\textbf{0.00000} & \cellcolor{gray!25}\textbf{0.03871} & \cellcolor{gray!25}\textbf{0.01950} & \cellcolor{gray!25}\textbf{0.00013} & \cellcolor{gray!25}\textbf{0.23559} & \cellcolor{gray!25}\textbf{0.13348} & \cellcolor{gray!25}\textbf{0.00043} & \cellcolor{gray!25}\textbf{0.38029} & \cellcolor{gray!25}\textbf{0.21126} & 0.00230 & 0.56234 & \cellcolor{gray!25}\textbf{0.32395} \\
& PINO     & 0.00001 & 0.06314 & 0.03970 & 0.00046 & 0.44367 & 0.32456 & 0.00134 & 0.66441 & 0.54963 & 0.00702 & 0.97468 & 0.69568 \\
& WNO      & 0.00002 & 0.07973 & 0.05021 & 0.00105 & 0.67300 & 0.44614 & 0.00451 & 1.22766 & 0.83277 & 0.01879 & 1.60840 & 0.99276 \\
& FINO     & 0.00002 & 0.07675 & 0.05457 & 0.00077 & 0.57917 & 0.49839 & 0.00204 & 0.82392 & 0.96417 & 0.00481 & 0.80971 & 1.14582 \\
\midrule
\multirow{7}{*}{E3}
& TF-SNO   & \cellcolor{gray!25}\textbf{0.00000} & \cellcolor{gray!25}\textbf{0.00783} & 0.01577 & \cellcolor{gray!25}\textbf{0.00003} & \cellcolor{gray!25}\textbf{0.01946} & \cellcolor{gray!25}\textbf{0.02556} & \cellcolor{gray!25}\textbf{0.00020} & \cellcolor{gray!25}\textbf{0.04326} & \cellcolor{gray!25}\textbf{0.04843} & \cellcolor{gray!25}\textbf{0.00090} & \cellcolor{gray!25}\textbf{0.08360} & \cellcolor{gray!25}\textbf{0.09108} \\
& FNO      & 0.00001 & 0.01274 & 0.01815 & 0.00055 & 0.07557 & 0.07950 & 0.00278 & 0.16599 & 0.17206 & 0.00897 & 0.29052 & 0.30053 \\
& DeepONet & 0.00486 & 0.25643 & 0.29367 & 0.11707 & 1.47394 & 1.53617 & 0.15451 & 1.62453 & 1.73887 & 0.27236 & 1.99208 & 2.25533 \\
& U-NO     & 0.00000 & 0.00835 & 0.01586 & 0.00009 & 0.03378 & 0.03844 & 0.00047 & 0.07514 & 0.08207 & 0.00156 & 0.13071 & 0.14055 \\
& PINO     & 0.00001 & 0.01030 & \cellcolor{gray!25}\textbf{0.01576} & 0.00029 & 0.06027 & 0.06407 & 0.00163 & 0.13549 & 0.14190 & 0.00657 & 0.25217 & 0.26599 \\
& WNO      & 0.00000 & 0.00921 & 0.01734 & 0.00020 & 0.04173 & 0.04936 & 0.00142 & 0.10455 & 0.11677 & 0.00731 & 0.21822 & 0.24741 \\
& FINO     & 0.00000 & 0.00816 & 0.01632 & 0.00004 & 0.02196 & 0.03023 & 0.00025 & 0.04849 & 0.05718 & 0.00113 & 0.09723 & 0.11015 \\
\midrule
\multirow{7}{*}{E4}
& TF-SNO   & \cellcolor{gray!25}\textbf{0.00000} & \cellcolor{gray!25}\textbf{0.00081} & \cellcolor{gray!25}\textbf{0.00164} & \cellcolor{gray!25}\textbf{0.00002} & \cellcolor{gray!25}\textbf{0.00639} & \cellcolor{gray!25}\textbf{0.01217} & \cellcolor{gray!25}\textbf{0.00009} & \cellcolor{gray!25}\textbf{0.01453} & \cellcolor{gray!25}\textbf{0.02747} & \cellcolor{gray!25}\textbf{0.00047} & \cellcolor{gray!25}\textbf{0.03245} & \cellcolor{gray!25}\textbf{0.05922} \\
& FNO      & 0.00000 & 0.00184 & 0.00361 & 0.00010 & 0.01499 & 0.02890 & 0.00045 & 0.03270 & 0.06090 & 0.00353 & 0.08669 & 0.17023 \\
& DeepONet & 0.00307 & 0.08022 & 0.16569 & 0.00105 & 0.04629 & 0.09435 & 0.00293 & 0.07853 & 0.13353 & 0.02947 & 0.23657 & 0.44270 \\
& U-NO     & 0.00000 & 0.00124 & 0.00235 & 0.00005 & 0.01016 & 0.01918 & 0.00025 & 0.02346 & 0.04339 & 0.00163 & 0.05584 & 0.10862 \\
& PINO     & 0.00000 & 0.00199 & 0.00367 & 0.00012 & 0.01650 & 0.02795 & 0.00059 & 0.03717 & 0.05946 & 0.00303 & 0.08284 & 0.14294 \\
& WNO      & 0.00002 & 0.00622 & 0.01063 & 0.00159 & 0.05982 & 0.09663 & 0.04654 & 0.28359 & 0.55953 & 2.55314 & 2.43995 & 4.72882 \\
& FINO     & 0.00000 & 0.00259 & 0.00429 & 0.00022 & 0.02278 & 0.03382 & 0.00119 & 0.05311 & 0.07361 & 0.00731 & 0.12744 & 0.21679 \\
\midrule
\multirow{7}{*}{E5}
& TF-SNO   & \cellcolor{gray!25}\textbf{0.00000} & \cellcolor{gray!25}\textbf{0.00187} & \cellcolor{gray!25}\textbf{0.00340} & 0.00011 & 0.01559 & \cellcolor{gray!25}\textbf{0.02800} & 0.00055 & 0.03558 & \cellcolor{gray!25}\textbf{0.06502} & \cellcolor{gray!25}\textbf{0.00254} & \cellcolor{gray!25}\textbf{0.07503} & \cellcolor{gray!25}\textbf{0.13571} \\
& FNO      & 0.00000 & 0.00304 & 0.00542 & 0.00027 & 0.02398 & 0.04245 & 0.00115 & 0.04999 & 0.08919 & 0.00431 & 0.09588 & 0.18417 \\
& DeepONet & 0.01249 & 0.16888 & 0.31414 & 0.10848 & 0.49503 & 0.62355 & 0.30561 & 0.81777 & 0.95512 & 0.31747 & 0.84556 & 1.02710 \\
& U-NO     & 0.00000 & 0.00188 & 0.00407 & \cellcolor{gray!25}\textbf{0.00010} & \cellcolor{gray!25}\textbf{0.01485} & 0.03058 & \cellcolor{gray!25}\textbf{0.00051} & \cellcolor{gray!25}\textbf{0.03354} & 0.06794 & 0.00437 & 0.09210 & 0.18944 \\
& PINO     & 0.00001 & 0.00569 & 0.00826 & 0.00110 & 0.04917 & 0.07160 & 0.00502 & 0.10604 & 0.15860 & 0.01768 & 0.19745 & 0.32905 \\
& WNO      & 0.00004 & 0.00985 & 0.01463 & 0.00446 & 0.09978 & 0.15861 & 0.04199 & 0.31476 & 0.74384 & 0.31628 & 0.86395 & 2.15247 \\
& FINO     & 0.00000 & 0.00256 & 0.00545 & 0.00022 & 0.02197 & 0.04293 & 0.00114 & 0.05070 & 0.08987 & 0.00500 & 0.10355 & 0.20213 \\
\midrule
\multirow{7}{*}{E6}
& TF-SNO   & \cellcolor{gray!25}\textbf{0.00000} & \cellcolor{gray!25}\textbf{0.00102} & \cellcolor{gray!25}\textbf{0.00264} & \cellcolor{gray!25}\textbf{0.00001} & \cellcolor{gray!25}\textbf{0.00727} & \cellcolor{gray!25}\textbf{0.01718} & \cellcolor{gray!25}\textbf{0.00002} & \cellcolor{gray!25}\textbf{0.01230} & \cellcolor{gray!25}\textbf{0.03009} & \cellcolor{gray!25}\textbf{0.00006} & \cellcolor{gray!25}\textbf{0.02107} & \cellcolor{gray!25}\textbf{0.05540} \\
& FNO      & 0.00000 & 0.00359 & 0.00605 & 0.00015 & 0.03116 & 0.05097 & 0.00076 & 0.07398 & 0.12257 & 0.00379 & 0.16228 & 0.28795 \\
& DeepONet & 0.00015 & 0.03273 & 0.09483 & 0.02070 & 0.34030 & 0.48508 & 0.09473 & 0.89076 & 1.27192 & 0.18043 & 1.37456 & 2.22763 \\
& U-NO     & 0.00000 & 0.00348 & 0.00372 & 0.00011 & 0.03310 & 0.02865 & 0.00067 & 0.08326 & 0.06595 & 0.00303 & 0.17935 & 0.14836 \\
& PINO     & 0.00000 & 0.00252 & 0.00631 & 0.00006 & 0.02192 & 0.05151 & 0.00044 & 0.05986 & 0.13265 & 0.00507 & 0.21627 & 0.45696 \\
& WNO      & 0.00000 & 0.00496 & 0.00800 & 0.00030 & 0.04912 & 0.07958 & 0.00178 & 0.12353 & 0.19999 & 0.00767 & 0.26656 & 0.43681 \\
& FINO     & 0.00002 & 0.01123 & 0.00620 & 0.00083 & 0.09293 & 0.05086 & 0.00587 & 0.25184 & 0.15816 & 0.02473 & 0.51456 & 0.48300 \\
\bottomrule
\end{tabular}
\end{table*}

\begin{table}[t]
\centering
\caption{Overall rollout results of all models on six non-stationary PDE benchmarks, in terms of MSE, H1, and L2.}
\label{tab:rollout_multistep}
\setlength{\tabcolsep}{2pt}
\resizebox{\linewidth}{!}{
\begin{tabular}{llccccccc}
\toprule
Eq. & Metric & TF-SNO & FNO & DeepONet & U-NO & PINO & WNO & FINO \\
\midrule
\multirow{3}{*}{E1}
& MSE & \cellcolor{gray!25}\textbf{0.00023} & 0.00264 & 0.00221 & 0.00030 & 0.03989 & 0.00411 & 0.00071 \\
& L2  & \cellcolor{gray!25}\textbf{0.03160} & 0.10174 & 0.09109 & 0.03827 & 0.42017 & 0.12894 & 0.05185 \\
& H1  & 0.04804 & 0.11722 & 0.17005 & \cellcolor{gray!25}\textbf{0.04310} & 0.50619 & 0.15696 & 0.06714 \\
\midrule
\multirow{3}{*}{E2}
& MSE & \cellcolor{gray!25}\textbf{0.02494} & 0.30122 & 0.26492 & 0.09992 & 0.21797 & 0.41718 & 0.08067 \\
& L2  & \cellcolor{gray!25}\textbf{0.36595} & 1.14827 & 2.19971 & 0.63305 & 1.05461 & 2.20815 & 0.72697 \\
& H1  & \cellcolor{gray!25}\textbf{0.32228} & 0.98157 & 1.62880 & 0.49130 & 0.94134 & 1.02960 & 0.78908 \\
\midrule
\multirow{3}{*}{E3}
& MSE & \cellcolor{gray!25}\textbf{0.01020} & 0.09366 & 0.25270 & 0.02480 & 0.11045 & 0.14545 & 0.01386 \\
& L2  & \cellcolor{gray!25}\textbf{0.16201} & 0.57844 & 1.49226 & 0.23788 & 0.44531 & 0.55838 & 0.19798 \\
& H1  & \cellcolor{gray!25}\textbf{0.17480} & 0.60039 & 1.54126 & 0.25156 & 0.47462 & 0.61629 & 0.21956 \\
\midrule
\multirow{3}{*}{E4}
& MSE & \cellcolor{gray!25}\textbf{0.00015} & 0.00102 & 0.01845 & 0.00077 & 0.00152 & 0.29952 & 0.00183 \\
& L2  & \cellcolor{gray!25}\textbf{0.01619} & 0.04096 & 0.16025 & 0.03525 & 0.04754 & 0.57238 & 0.05565 \\
& H1  & \cellcolor{gray!25}\textbf{0.02845} & 0.07078 & 0.28976 & 0.06223 & 0.08034 & 1.21476 & 0.08426 \\
\midrule
\multirow{3}{*}{E5}
& MSE & \cellcolor{gray!25}\textbf{0.00067} & 0.00144 & 0.26797 & 0.00105 & 0.00501 & 0.08463 & 0.00161 \\
& L2  & \cellcolor{gray!25}\textbf{0.03366} & 0.04872 & 0.72193 & 0.03917 & 0.09430 & 0.35866 & 0.05150 \\
& H1  & \cellcolor{gray!25}\textbf{0.06318} & 0.08941 & 0.85333 & 0.07912 & 0.14566 & 0.91347 & 0.09725 \\
\midrule
\multirow{3}{*}{E6}
& MSE & \cellcolor{gray!25}\textbf{0.00002} & 0.00113 & 0.10747 & 0.00096 & 0.00105 & 0.00246 & 0.00803 \\
& L2  & \cellcolor{gray!25}\textbf{0.01185} & 0.07680 & 0.88314 & 0.08568 & 0.07548 & 0.12670 & 0.25175 \\
& H1  & \cellcolor{gray!25}\textbf{0.02953} & 0.13018 & 1.34664 & 0.06937 & 0.16342 & 0.20547 & 0.18533 \\
\bottomrule
\end{tabular}
}
\end{table}

\begin{table}[t]
\centering
\small
\caption{Ablation results on E1 and E4 removing different components of TF-SNO.}
\label{tab:ablation}
\setlength{\tabcolsep}{6pt}
\begin{tabular}{llccc}
\toprule
Equation & Variant & MSE & L2 & H1 \\ \midrule
\multirow{6}{*}{\makecell[l]{E1: 1D\\ Burgers}} & Full TF-SNO & \cellcolor{gray!25}\textbf{0.00023} & \cellcolor{gray!25}\textbf{0.03160} & \cellcolor{gray!25}\textbf{0.04804} \\
   & w/o $\delta_\theta$ & 0.00052 & 0.04791 & 0.06963 \\
   & Scalar $\delta_\theta$ & 0.00035 & 0.03711 & 0.05038 \\
   & w/o $s_{\mathrm{time}}$ & 0.00044 & 0.04379 & 0.06389 \\
   & Energy Statistic Only & 0.00074 & 0.05824 & 0.05511 \\
   & w/o H$^1$ Loss & 0.00063 & 0.05277 & 0.07403 \\
\midrule
\multirow{6}{*}{\makecell[l]{E4: 2D\\ Burgers}} & Full TF-SNO & \cellcolor{gray!25}\textbf{0.00015} & \cellcolor{gray!25}\textbf{0.01619} & \cellcolor{gray!25}\textbf{0.02845} \\
   & w/o $\delta_\theta$ & 0.00176 & 0.05474 & 0.11166 \\
   & Scalar $\delta_\theta$ & 0.00180 & 0.05361 & 0.10914 \\
   & w/o $s_{\mathrm{time}}$ & 0.00129 & 0.04729 & 0.09538 \\
   & Energy Statistic Only & 0.00163 & 0.05187 & 0.10676 \\
   & w/o H$^1$ Loss & 0.00249 & 0.06189 & 0.13349 \\
\bottomrule
\end{tabular}
\end{table}

In this section, we conduct a comprehensive experimental study to evaluate TF-SNO, including the problem settings, implementation details, comparison results, and experimental analysis.

\subsection{Experimental Setup}

\subsubsection{Benchmarks}
To evaluate \textit{TF-SNO} on non-stationary dynamics with evolving spectra and multiscale behavior, we consider six PDEs from PDEBench \cite{takamoto22_neurips} covering nonlinear convection-diffusion, reaction-diffusion with phase separation, chaotic high-order dynamics, and multiscale vortex interaction, including the \textit{1D Burgers equation (E1), 1D Allen-Cahn equation (E2), 1D Kuramoto-Sivashinsky (KS) equation (E3), 2D scalar Burgers equation (E4), 2D Allen-Cahn equation (E5)}, and \textit{2D Navier-Stokes Equation (E6)}. We use the same PDE parameter settings for both training and testing, except for cross-parameter experiments. 

\subsubsection{Baselines}
We compare TF-SNO with six representative operator-learning baselines. \textit{Fourier Neural Operator (FNO)} \cite{li2021fourier} is a global FFT-based spectral neural operator serving as a basis model. \textit{DeepONet} \cite{lu2021learning} is a coordinate-conditioned operator regression framework with universal approximation. \textit{U-Shaped Neural Operator (U-NO)} \cite{rahman2023uno} introduces a U-shaped encoder-decoder backbone for multiscale feature fusion in neural operators. \textit{Physics-Informed Neural Operator (PINO)} \cite{li2024physics} augments operator learning with PDE-residual physics constraints. \textit{Wavelet Neural Operator (WNO)} \cite{tripura2023wavelet} uses wavelet multi-resolution representations to better capture localized transients. \textit{FINO} \cite{cheng2025pde} is a finite-difference-inspired learned solver that emphasizes local stencil-based updates for fast and stable long-horizon rollouts.

\subsubsection{Evaluation Metrics}
We report three complementary metrics to evaluate accuracy and rollout quality.
\textit{Mean squared error (MSE)} measures pointwise prediction accuracy on the discrete grid by averaging the squared difference between the predicted and ground-truth fields over all grid points and output channels. For a rollout of length $T$, we report $\mathrm{MSE}_{\mathrm{roll}} = \frac{1}{T}\sum_{t=1}^{T}\left(\frac{1}{N}\sum_{i=1}^{N}\left(\hat{u}_i^{t}-u_i^{t}\right)^2\right)$.
\textit{Relative L$_2$ error (L2)} measures the normalized field-value error: $\varepsilon_{L^2}^{\mathrm{rel}} = \frac{\|\hat{u}-u\|_{L^2(\Omega)}}{\|u\|_{L^2(\Omega)}+\epsilon} \approx \sqrt{\frac{\sum_{i=1}^N \|\hat{u}_i - u_i\|^2}{\sum_{i=1}^N \|u_i\|^2+\epsilon}}$, where $\epsilon$ is a small constant for numerical stability. 
\textit{Relative H$^1$ seminorm error (H1)} evaluates gradient-level accuracy. We first compute the spatial gradients of the prediction and target using the same finite-difference operator as in training, and then normalize the gradient error by the target-gradient norm: $\varepsilon_{H^1}^{\mathrm{rel}}=\frac{\|\nabla_h (\hat{u}-u)\|_{L^2(\Omega)}}{\|\nabla_h  u\|_{L^2(\Omega)}}\approx \sqrt{\frac{\sum_{i=1}^{N} \|\nabla_h  \hat{u}_i-\nabla_h  u_i\|^2}{\sum_{i=1}^{N} \|\nabla_h  u_i\|^2}}$. For 2D equations, $\nabla_h$ includes derivatives along both spatial directions. Lower values indicate better performance for all metrics.

\subsubsection{Implementation Details}
All methods are implemented in PyTorch and run on a single NVIDIA RTX 4090 (24GB). For each benchmark, we use 1,000 training trajectories and 200 independent test trajectories. Models are trained at fixed resolutions to enforce invariance, 256 for 1D PDEs and $64{\times}64$ for 2D PDEs, with inputs downsampled during training and resampled back to the original resolution at inference.
We use AdamW optimizer \cite{loshchilov2017decoupled} with initial learning rate $10^{-3}$, weight decay $10^{-4}$, cosine annealing \cite{loshchilov2017sgdr} to $10^{-5}$, 500 training epochs, and pushforward horizon $T{=}5$ unless otherwise specified.

\subsection{Main Results}

Table~\ref{tab:rollout_horizons} reports step-wise rollout errors at horizons \{1,10,25,50\} on all six benchmarks.
TF-SNO maintains the most stable error growth across horizons. It performs best on E1 and E3, becomes competitive on E2 at Step=50, and achieves the best or near-best results on 2D tasks, where several baselines show clear drift or instability, such as WNO on E4/E5 and DeepONet on E5/E6. These results indicate that state-adaptive time-frequency modulation reduces long-horizon error accumulation under evolving spectra.

Table \ref{tab:rollout_multistep} summarizes overall rollout performance. Across E1-E6, TF-SNO consistently obtains the lowest MSE and $L^2$ error, with larger gains on the harder 2D benchmarks. For the strongest baseline, U-NO achieves the best $H^1$ on E1, but TF-SNO remains better on the other tasks. Overall, the results show that adaptive band-wise spectral routing improves both field-level accuracy and gradient-level stability for non-stationary PDE rollouts.

\begin{table}[t]
\centering
\caption{Group-wise and single-feature ablations for the 8D controller. Values are Step=50 relative $L^2$ errors.}
\label{tab:feature}
\begin{tabular}{lcc|lcc}
\toprule
Config. & E3 & E6 & Config. & E3 & E6 \\
\midrule
w/o $s_{\mathrm{time}}$ & 0.0978 & 0.0249 & w/o $s_{\mathrm{freq}}$ & 0.1046 & 0.0268 \\
w/o $\log\sigma_z$ & 0.0867 & 0.0222 & w/o $\log E$ & 0.0897 & 0.0244 \\
w/o $R_{\mathrm{rough}}$ & 0.0902 & 0.0233 & w/o $\mu_{\mathcal{K}}$ & 0.0925 & 0.0252 \\
w/o $\kappa_z$ & 0.0873 & 0.0226 & w/o $\sigma_{\mathcal{K}}$ & 0.0916 & 0.0260 \\
w/o $c$ & 0.0888 & 0.0230 & w/o $\mathrm{Skew}_{\mathcal{K}}$ & 0.0904 & 0.0237 \\
\rowcolor{gray!25}
\textbf{Full} & \textbf{0.0836} & \textbf{0.0211} & \textbf{Full} & \textbf{0.0836} & \textbf{0.0211} \\
\bottomrule
\end{tabular}
\end{table}

\begin{table}[t]
\centering
\small
\caption{Cross-Reynolds generalization on E6. Models are trained on the source domain ($Re=500$) and evaluated on both the source and target domain ($Re=1000$).}
\label{tab:cross_re}
\setlength{\tabcolsep}{2.5pt}
\begin{tabular}{l|ccc|ccc}
\toprule
\multirow{2}{*}{Model} & \multicolumn{3}{c|}{$Re=500$ (Source)} & \multicolumn{3}{c}{$Re=1000$ (Target)} \\
\cmidrule(lr){2-4}\cmidrule(lr){5-7}
 & MSE & L2 & H1 & MSE & L2 & H1 \\
\midrule
TF-SNO (Full)                 & \cellcolor{gray!25}\textbf{0.00208} & \cellcolor{gray!25}\textbf{0.03772} & \cellcolor{gray!25}\textbf{0.05037} & \cellcolor{gray!25}\textbf{0.00062} & \cellcolor{gray!25}\textbf{0.05422} & \cellcolor{gray!25}\textbf{0.07349} \\
w/o gating                    & 0.00439 & 0.06615 & 0.08970 & 0.00248 & 0.08871 & 0.12876 \\
w/o multiscale                & 0.00298 & 0.05344 & 0.12361 & 0.00130 & 0.07070 & 0.09769 \\
FNO                           & 0.02719 & 0.15981 & 0.19032 & 0.00821 & 0.15613 & 0.15863 \\
\bottomrule
\end{tabular}
\end{table}

\begin{figure*}[ht]
  \centering
  \includegraphics[width=0.8\linewidth]{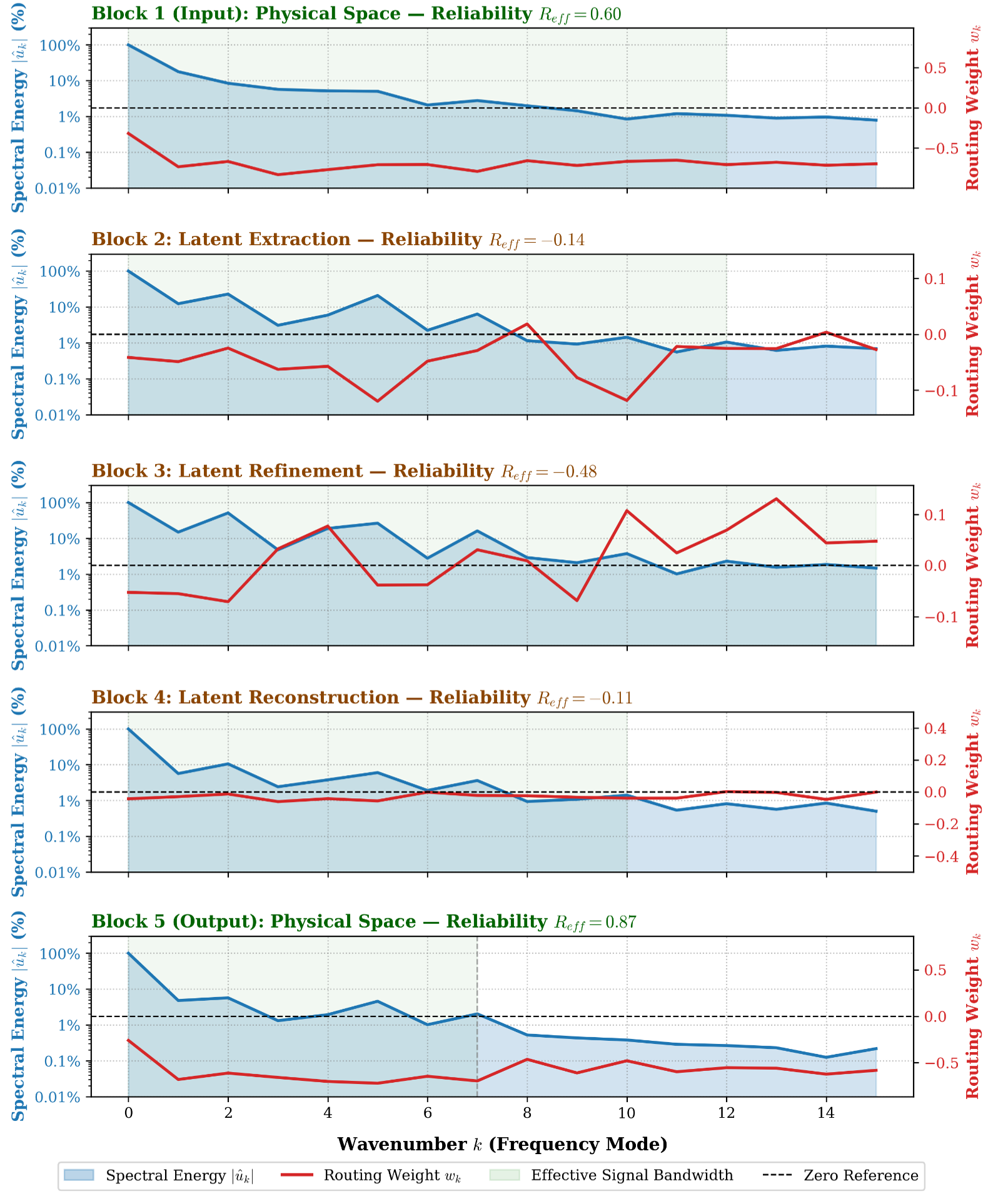}
  \caption{Calibration of frequency-band energy in TF-SNO model. Blue lines with left axis represent spectrum energy, red lines with right axis represent learned band routing weights, and green regions show the effective bandwidths.}
  \label{fig:band}
\end{figure*}

\begin{table}[t]
\centering
\caption{Comparison with richer controller statistics alternatives. Values are Step=50 relative $L^2$ errors.}
\label{tab:controller}
\begin{tabular}{lccc}
\toprule
Model & E3 & E5 & E6 \\
\midrule
TF-SNO (8D statistics) & \cellcolor{gray!25}\textbf{0.0836} & \cellcolor{gray!25}\textbf{0.0750} & 0.0211 \\
16D direction-aware controller & 0.0912 & 0.0814 & 0.0196 \\
32D learned controller & 0.0965 & 0.0789 & \cellcolor{gray!25}\textbf{0.0187} \\
\bottomrule
\end{tabular}
\end{table}

\begin{table}[t]
\centering
\caption{Spectral-level metrics at the rollout endpoint.}
\label{tab:spectral}
\begin{tabular}{llcc}
\toprule
Benchmark & Model & PSRE$\downarrow$ & SCE$\downarrow$ \\
\midrule
\multirow{3}{*}{E3} & FNO & 0.453 & 12.54 \\
& U-NO & 0.282 & 8.26 \\
& TF-SNO & \cellcolor{gray!25}\textbf{0.153} & \cellcolor{gray!25}\textbf{3.15} \\
\midrule
\multirow{3}{*}{E6} & FNO & 0.251 & 6.42 \\
& U-NO & 0.128 & 3.85 \\
& TF-SNO & \cellcolor{gray!25}\textbf{0.045} & \cellcolor{gray!25}\textbf{1.21} \\
\bottomrule
\end{tabular}
\end{table}

\subsection{Ablation Studies}

\subsubsection{Framework Component Ablation}
We evaluate the contribution of TF-SNO components on both 1D (E1) and 2D (E4) benchmarks. 
The full model uses band-wise state-dependent gating $\delta_\theta(z)$ driven by joint time-frequency statistics $s_{\mathrm{tf}}=[s_{\mathrm{freq}},s_{\mathrm{time}}]$, and is trained with an L$_2$ loss plus a global H$^1$ seminorm. 
As shown in Table~\ref{tab:ablation}, the full TF-SNO performs best on both PDEs. Removing state-dependent gating $\delta_\theta$ causes the most severe degradation, particularly on E4, while replacing it with a scalar gate recovers part of the loss but remains clearly inferior, indicating the necessity of band-wise adaptation. Reducing the gating input to weaker statistics, such as disabling $s_{\mathrm{time}}$ or even using only the energy statistic in $s_{\mathrm{freq}}$, consistently harms accuracy, and removing the H$^1$ loss further increases error, supporting the role of gradient-level supervision in stabilizing rollouts. 
We further conduct group-wise and single-feature ablations on E3 and E6 in Table \ref{tab:feature}. Removing either group degrades performance, and removing individual statistics also increases error.

\subsubsection{Cross-Parameter Generalization}

We conduct a cross-parameter (Reynolds) generalization experiment on the E6 benchmark to evaluate whether TF-SNO can transfer learned dynamics across viscosity regimes. As shown in Table~\ref{tab:cross_re}, TF-SNO achieves the lowest errors on both the source domain ($Re=500$) and the target domain ($Re=1000$), indicating that the proposed state-adaptive modulation remains effective under cross-parameter shifts. 
The lower MSE on $Re=1000$ is mainly because MSE is computed as a time-averaged error over all rollout steps, and the faster energy decay in later stages of the $Re=1000$ trajectories drives the overall field magnitude closer to zero, reducing the absolute error scale.

\subsubsection{Sensitivity to Gating Signals}

To address whether the compact 8D statistics are too restrictive, we compare TF-SNO with richer controllers. We first expand the statistics from 8D to 16D by adding directional spectral centroid and quadrant-wise band energy. And we replace handcrafted statistics with a lightweight convolutional encoder that produces a 32D latent signal.
As shown in Table \ref{tab:controller}, extended controllers improve performance on more anisotropic E6 but increase error on E5 and E3, suggesting that they are not uniformly beneficial across PDE types. We therefore characterize the 8D design not as universally optimal, but as a task-agnostic efficiency-accuracy trade-off that is lightweight, interpretable, and stable across PDE types.

We also evaluate MetaNet design choices on E6 due to its complex spectral dynamics. The default two-layer MLP with \texttt{width=32} and ReLU achieves Step=50 $L^2$ error $0.02107\pm0.00162$. A minimal one-layer linear network with \texttt{width=16} increases error to $0.02340\pm0.00186$, confirming that moderate nonlinear capacity is useful. A larger three-layer MLP with \texttt{width=64} gives $0.02182\pm0.00170$, showing no consistent gain. GELU and SiLU activations give $0.02091\pm0.00155$ and $0.02143\pm0.00166$, respectively. Initializing $\alpha$ at $0.1$ and $0.5$ gives $0.02174\pm0.00173$ and $0.02120\pm0.00160$. These results indicate that TF-SNO is relatively insensitive to common MetaNet choices within a reasonable capacity range.

\subsection{Band-Energy Calibration and Spectral-Level Validation}

Fig. \ref{fig:band} reports a band-energy calibration analysis of TF-SNO by comparing the physical power spectrum with the learned gating weights across Fourier wavenumbers. We quantify reliability using the correlation within an effective bandwidth, defined as modes whose spectral energy exceeds $1\%$ of the maximum energy. The results suggest a consistent hierarchical processing pattern across blocks. At the physical-space interface and final reconstruction stage, the model exhibits positive alignment between dominant low-frequency energy and routing strength, with $R_{\mathrm{eff}}=0.60$ and $0.87$, respectively. In contrast, intermediate latent blocks show weak or negative correlations, with $R_{\mathrm{eff}}=-0.14$, $-0.48$, and $-0.11$, implying that the gating mechanism reallocates capacity toward low-energy components with higher-frequency residual structures that are not dominant in the global spectrum but are crucial for sharp fronts and transient bursts.

To further quantify spectral behavior, we report two endpoint spectral metrics on E3 and E6, which best reflect complex, non-stationary spectral behavior. 
Power Spectrum Relative Error $\mathrm{PSRE}=\frac{\|P_{pred}-P_{gt}\|_2}{\|P_{gt}\|_2}$ measures overall energy distribution fidelity, and
Spectral Centroid Error $\mathrm{SCE}=|\mu_{pred}-\mu_{gt}|$ measures dominant frequency, where $\mu=\frac{\sum_k \|k\|\,P(k)}{\sum_k P(k)}$. Table \ref{tab:spectral} shows that TF-SNO improves both metrics over FNO and U-NO, indicating that the gains extend beyond physical-space rollout accuracy to more faithful tracking of spectral energy distribution and dominant frequency.

\section{Conclusion}
In this paper, we propose TF-SNO, a state-adaptive spectral neural operator that performs band-wise time-frequency modulation driven by compact statistics of the evolving state, and integrates these adaptive spectral blocks into a multiscale U-shaped backbone to jointly model global structures and transient local dynamics in non-stationary PDEs. This framework provides an explicit mechanism to adjust spectral responses across phases of evolution while preserving FFT-based efficiency, with H$^1$ regularization constraints to improve gradient-level fidelity and rollout stability.
Experiments across six benchmarks validate the effectiveness of TF-SNO, with ablations indicating the contribution of dynamic gating and statistics. Future work may include extending the framework to irregular geometries and mesh-based settings, developing stability-aware objectives for longer rollouts, and exploring tighter couplings between global spectral updates and learned local stencil dynamics.




\begin{acks}
This work was supported by the National Natural Science Foundation of China (52406079) and Shenzhen Municipal Science and Technology Innovation Bureau (JCYJ20250604145646062).
\end{acks}

\bibliographystyle{ACM-Reference-Format}
\balance
\bibliography{ref}


\end{document}